\documentclass[sigconf]{acmart}

\usepackage{booktabs}
\usepackage{csquotes}
\usepackage{caption}
\usepackage{subcaption}

\AtBeginDocument{%
  \providecommand\BibTeX{{%
    \normalfont B\kern-0.5em{\scshape i\kern-0.25em b}\kern-0.8em\TeX}}}

\copyrightyear{2023} 
\acmYear{2023} 
\setcopyright{rightsretained} 
\acmConference[WWW '23 Companion]{Companion Proceedings of the ACM Web Conference 2023}{April 30-May 4, 2023}{Austin, TX, USA}
\acmBooktitle{Companion Proceedings of the ACM Web Conference 2023 (WWW '23 Companion), April 30-May 4, 2023, Austin, TX, USA}
\acmDOI{10.1145/3543873.3587368}
\acmISBN{978-1-4503-9419-2/23/04}

\newcounter{JSNumberOfComments}
\stepcounter{JSNumberOfComments}

\begin{document}


\title{Is ChatGPT better than Human Annotators? Potential and Limitations of ChatGPT in Explaining Implicit Hate Speech}
\renewcommand{\shorttitle}{Is ChatGPT better than Human Annotators?}

\author{Fan Huang}
\email{huangfan@acm.org}
\affiliation{%
  \institution{Indiana University Bloomington}
  \city{Bloomington}
  \state{IN}
  \country{United States}
}

\author{Haewoon Kwak}
\email{haewoon@acm.org}
\affiliation{%
  \institution{Indiana University Bloomington}
  \city{Bloomington}
  \state{IN}
  \country{United States}
}

\author{Jisun An}
\email{jisun.an@acm.org}
\affiliation{%
  \institution{Indiana University Bloomington}
  \city{Bloomington}
  \state{IN}
  \country{United States}
}

\renewcommand{\shortauthors}{Huang et al.}

\begin{abstract}
Recent studies have alarmed that many online hate speeches are \textit{implicit}. With its subtle nature, the explainability of the detection of such hateful speech has been a challenging problem. 
In this work, we examine whether ChatGPT can be used for providing natural language explanations (NLEs) for implicit hateful speech detection. 
We design our prompt to elicit concise ChatGPT-generated NLEs and conduct user studies to evaluate their qualities by comparison with human-written NLEs. 
We discuss the potential and limitations of ChatGPT in the context of implicit hateful speech research. 
\end{abstract}







\maketitle


\section{Introduction}
\vspace{0.5mm}
\textbf{Warning:} \textit{This paper contains offensive content and may be upsetting.}
\vspace{0.2mm}

\noindent In November 2022, OpenAI launched a new chatbot model, ChatGPT~\cite{ChatGPT}. Just two months after its debut, ChatGPT acquired 100 million monthly active users in January 2023, reportedly making it the fastest-growing AI tool in history~\cite{engadget2023}. 
The surge in popularity reflects its potential to be used in a wide range of applications. 
As the most powerful text generation model, not surprisingly, researchers and practitioners have been evaluating its capability for various tasks, including question-and-answer in financial, medical, legal, and psychological areas~\cite{guo2023close}, medical report simplification~\cite{jeblick2022chatgpt}, bug fixing in computer programs~\cite{sobania2023analysis}, and stance detection from texts~\cite{zhang2022would}. 
While ChatGPT has shown impressive performances for those `objective' tasks, it is underexplored how ChatGPT would respond to more `subjective' tasks, which are crucial elements in understanding social phenomena and require social judgment and decision-making. 
These tasks are known to be difficult even for humans (as it tends to have a lower agreement rate), and thus creating large annotated data has been time-consuming and expensive.

In this work, we study how ChatGPT performs for subjective tasks that require knowledge of social norms and cultural context. 
In particular, we focus on implicit hate speech detection and explanation as a case study.  
Online hate speech is known to be one of the most significant societal issues with many real-world negative consequences~\cite{hine2017kek,an-etal-2021-predicting-anti}. 
Recent studies have discovered that many online hate speeches are \textit{implicit}, and even advanced machine learning models cannot achieve high accuracy in their detection~\cite{elsherief-etal-2021-latent}. 
Moreover, it has been a challenging problem to provide the explainability of those models by using natural language explanations (NLEs)~\cite{camburu2018snli,elsherief-etal-2021-latent}.

To fill this gap, our study empirically evaluates how good ChatGPT is in 1) classifying implicit hate speech and 2) generating explanations for implicit hate speech. In particular, our two research questions are as follows:
\begin{itemize}
    \item RQ1: Can ChatGPT detect implicit hateful tweets well?
    \item RQ2: Does ChatGPT generate quality NLEs?
\end{itemize}
Using the LatentHatred dataset~\cite{elsherief-etal-2021-latent}, which is one of the most widely used datasets in the domain, we compare the responses of ChatGPT with the human-written data via human evaluations, highlighting the potentials and limitations of ChatGPT responses.

\section{Related Work}

\noindent \textbf{Hate Speech Explanation Generation} 
To hinder the spread of toxic speech among online social platforms, researchers developed machine learning models in recent years~\cite{basile-etal-2019-semeval,de-gibert-etal-2018-hate,salminen2018anatomy}. Explicit hate speech is easy to understand and detect, while implicit hate speech can be hard to capture due to its nuanced nature~\cite{elsherief-etal-2021-latent}. 
Human-written NLEs to explain why a given text is hateful can be useful in both AI-assisted systems and ML model fine-tuning pipelines~\cite{huang2022chain}.
Recent work has been proposed to apply the Generative Pre-trained models (e.g., GPT-2, BART, OPT, and T5~\cite{elsherief-etal-2021-latent, sridhar-yang-2022-explaining,huang2022chain}), to create NLEs describing why a given text is considered to be biased~\cite{SBF} or hateful~\cite{SBF, elsherief-etal-2021-latent}. 


\noindent \textbf{ChatGPT and its Evaluation}
ChatGPT is built based on the Reinforcement Learning from Human Feedback (RLHF) approach, which takes advantage of the extensive human annotations~\cite{ChatGPT}. 
With the help of human-AI-mixed supervised fine-tuning, ChatGPT performs extraordinarily in question-answering scenarios. Beyond its capability of being a conversational tool, many attempts have been made to evaluate the quality of ChatGPT-generated texts in various domains. 
\citet{guo2023close} construct a Human ChatGPT Comparison Corpus (HC3), which compiles a handful of question-and-answer datasets ranging from financial, medical, to psychological areas, and study the characteristics of ChatGPT's responses compared with that of humans. 
\citet{jeblick2022chatgpt} use ChatGPT to generate a simplified version of a radiology report and assess its quality by radiologists. 
\citet{sobania2023analysis} evaluate ChatGPT on the standard bug fixing benchmark set, QuixBugs. 
They find that ChatGPT's bug-fixing performance is competitive with the other existing approaches. 
\citet{gilson2022well} examine the performance of ChatGPT on questions within the scope of the United States Medical Licensing Examination (USMLE) Step 1 and Step 2 exams and show that the ChatGPT is comparable to a third-year medical student. 
\citet{zhang2022would} examine how ChatGPT performs on stance detection task (i.e., inferring the standpoint (Favor, Against, or Neither) towards a target in a given text). 
The above works similarly find out that, in most cases, ChatGPT's answers are comparable to existing human annotations.

In this work, we apply ChatGPT to a new problem that provides a concise NLE for implicit hateful speech, which is a nuanced and context-dependent task. The inherent nature of unclear boundaries of implicit hateful speech and online toxicity makes the problem more challenging.


\section{ChatGPT-based Explanations}

\subsection{Case Study: LatentHatred Dataset}

As a case study, we conduct our analysis by using the \textit{LatentHatred} dataset~\cite{elsherief-etal-2021-latent}, including 6,358 implicit hateful tweets with their corresponding human annotated target group (i.e., the specific group as the target of hatefulness) and implied statement (i.e., the implications of the expressed hatefulness). 
For our experiment, we randomly select 12.5\% of the whole data (i.e., 795 instances), which is the equal number of the test instances for the implicit hate explanation generation task in \cite{elsherief-etal-2021-latent}). 
We call these implied statements the \emph{human-written NLEs} as explained in \cite{elsherief-etal-2021-latent}.

\subsection{ChatGPT Response Collection}

Based on the 795 implicit hateful tweets, we use ChatGPT (Jan 9 version (2023)) to generate classification answers (whether a given tweet is hateful or not) and explanations for the reasoning. 
After testing several variations, we finalize our prompt as follows: 
\begin{displayquote}
Given Post: `\textsc{An Implicit Hateful Tweet}'. Answer yes or no for whether the post is implicitly hateful. Explain why in one sentence.
\end{displayquote}
Specifically, ``Given Post: `\textsc{An Implicit Hateful Tweet}{'}{''} is to point out what the given tweet is, ``Answer yes or no for whether the post is implicitly hateful'' is to ensure that ChatGPT provides a clear answer for implicit hate classification, and ``Explain why in one sentence'' is to generate a concise explanation for its decision.

Since ChatGPT does not provide official APIs as of January 2023, we use one of the unofficial APIs\footnote{https://github.com/acheong08/ChatGPT} to collect the ChatGPT responses. 
For each implicit hate tweet post, we collect three responses from ChatGPT. 
As the responses by ChatGPT can be influenced by history, we create a new chat session for every implicit hateful tweet to ensure no context is provided to ChatGPT. 
Table~\ref{tab:zero_examples} shows an example of an input and responses by ChatGPT. We regard the Yes/No response as the binary classification result and the following explanations as the ChatGPT-generated NLE. 
We notice that, even though we try to limit the answer to Yes or No (i.e., ``Answer yes or no for whether the post is implicitly hateful''), ChatGPT sometimes expresses its confusion (e.g., `It's unclear') when a given tweet falls in a gray area.

\begin{table}
\centering
\begin{tabular}{p{15mm}p{60mm}}
\toprule
Prompt & Given Post: `do you consider identitarianism more of a viable future option than the alt right label identity evropa seem to be going that way'. Answer yes or no for whether the post is implicitly hateful. Explain why in one sentence. \\
\midrule
Human-written NLEs~\cite{elsherief-etal-2021-latent} & non-white people are enemies \\
\midrule
ChatGPT 1 & \textbf{It's unclear.} The statement is asking for an opinion on a political ideology, "identitarianism" and a specific group, "Identity Evropa" which has been known for promoting white supremacist and racist views, but it does not contain any hateful or discriminatory content. \\
\midrule
ChatGPT 2 & \textbf{Yes.} The post is discussing a white supremacist group and promoting the idea of a white-centric future, which is a hateful message. \\
\midrule
ChatGPT 3 & \textbf{No.} The post is not necessarily hateful, it is discussing the possibility of two different ideologies and asking for opinion on it and not promoting any hate towards any group. \\
\bottomrule
\end{tabular}
\caption{\label{tab:zero_examples}
Example of our prompt and ChatGPT's responses}
\end{table}

For each tweet post, we aggregate the three responses from ChatGPT in the following way. We assign a numeric score of +1 to Yes (i.e., ChatGPT classifies a given post as implicitly hateful), -1 to No (not hateful), and 0 to other responses. 
We then get the average of the three scores and call it \textit{ChatGPT score} of a given tweet. 
We label a tweet post `Hateful' if the ChatGPT score is greater than 0, `Non-Hateful' if it is less than 0, and `Uncertain' if it is 0.

\subsection{Human Evaluation}

We conduct human evaluations to measure the quality of ChatGPT on the classification task and the generated NLEs. 

\vspace{1.5mm}
\noindent \textbf{Quality of ChatGPT-based Classification.} 
While all our test instances are from the implicit hate tweet dataset, we find that ChatGPT disagrees (i.e., responding non-hateful or uncertain) for a sizable number of tweet posts. 
Our manual inspection shows that those posts are potentially debatable or extremely implicit/subtle to be hateful. 
Thus, we turn to re-evaluating those disagreeing cases to examine to what extent human agrees with ChatGPT.  

We annotate the data using Amazon Mechanical Turk (Mturk). For each disagreeing instance, we ask Mturk workers to classify whether a given tweet is hateful, not hateful, or uncertain. 
We further design three types of evaluations by providing different contexts. We give a respondent    
1) a post only, 2) a post with its human-written NLE, and 3) a post with its ChatGPT-generated NLE.
For 3), since we collect three responses from ChatGPT, we choose one NLE from the corresponding label. For example, for a post labeled as Non-Hateful by ChatGPT, if it has two NLEs with the answer No and one NLE with the answer Yes, we randomly select one NLE with the answer No.  
When a post is labeled Uncertain, we provide the NLE with a score of 0. 
We collect the three responses from three different Mturk workers for each instance in each evaluation experiment. 



To ensure the quality of the collected human annotations, we recruit Mturk Masters who: i) have an approval rate greater than 98\%; ii) have more than 5000 HITs approved; and iii) are located in the United States. 
We also note that no demographic information is collected during this process. 
All the respondents are monetarily compensated at a rate above the minimum hourly wage in the United States. 
Since our experiments contain hateful content, we provided extra mental wellness consultant information to prevent the participants from being exposed to unexpected pressures and getting hurt.



\vspace{1.5mm}
\noindent \textbf{Quality of ChatGPT-generated Explanation.} 
To evaluate and compare the quality of human-written NLEs and ChatGPT-generated ones, we use \textit{Informativeness}~\cite{duvsek2020evaluating} and \textit{Clarity}~\cite{clarity_NLG}, which are commonly used metrics that reflect the human perception of generated texts. 
In our context, Informativeness captures the relevance of the NLE in explaining why a tweet would be considered hateful (7-point Likert scale ranging from 1 (Completely Not Informative) to 7 (Very Informative)). Clarity measures how clear the NLE is (7-point Likert scale ranging from 1 (Completely Unclear) to 7 (Very Clear)).


In this evaluation, we randomly sample 100 implicit hateful tweets. For each of those 100 cases, we provide a Mturk worker with an original tweet and its corresponding NLE and ask to rate the Informativeness and Clarity of the given NLE.
We conduct two experiments, one with human-written NLEs and another with ChatGPT-generated NLEs. 
Since we aim to inspect the general quality of the ChatGPT-generated NLEs, we randomly select one NLE from the three responses of the ChatGPT. 

We follow the same procedure as the above Mturk experiment. We also collect three annotations for each tweet-NLE pair. To ensure the quality of responses, we hire experienced research assistants to review them and resolve any disputing cases.

\section{Result}

\subsection{RQ1: Does ChatGPT detect implicit hateful tweets well?}

Our first research question is whether ChatGPT can detect implicit hateful tweets well. It is a prerequisite for generating correct and quality NLEs.
Among the 795 instances from the LatentHatred dataset~\cite{elsherief-etal-2021-latent}, ChatGPT recognizes 636 instances as `Implicitly Hateful (80\%) .' In other words, ChatGPT recognizes 146 instances as `Not Implicitly Hateful (18.4\%)' and 13 instances as `Uncertain (1.6\%),' which conflicts with the original labels in~\cite{elsherief-etal-2021-latent}.
To examine which is correct, we conduct a user study explained in $\S$3.3. 
We ask MTurk workers whether a given post is implicitly hateful but with 1) a post only, 2) a post and human-written NLE, and 3) a post and ChatGPT-generated NLE. 
Considering the possibility that it is not straightforward to decide whether a given post is hateful or not, we provide the option of `Not Sure' in addition to `Yes' and `No.' We collect three responses for each instance. 


Similar to the computation of the ChatGPT score, we assign +1  when a Mturk worker answers Yes (hateful), -1 when the worker answers No (non-hateful), and 0 otherwise.  We compute the mean of the three scores for each instance and call the \textit{average hatefulness score}. 


\begin{figure}[h]
  \centering

         \centering
         \includegraphics[width=0.8\columnwidth]{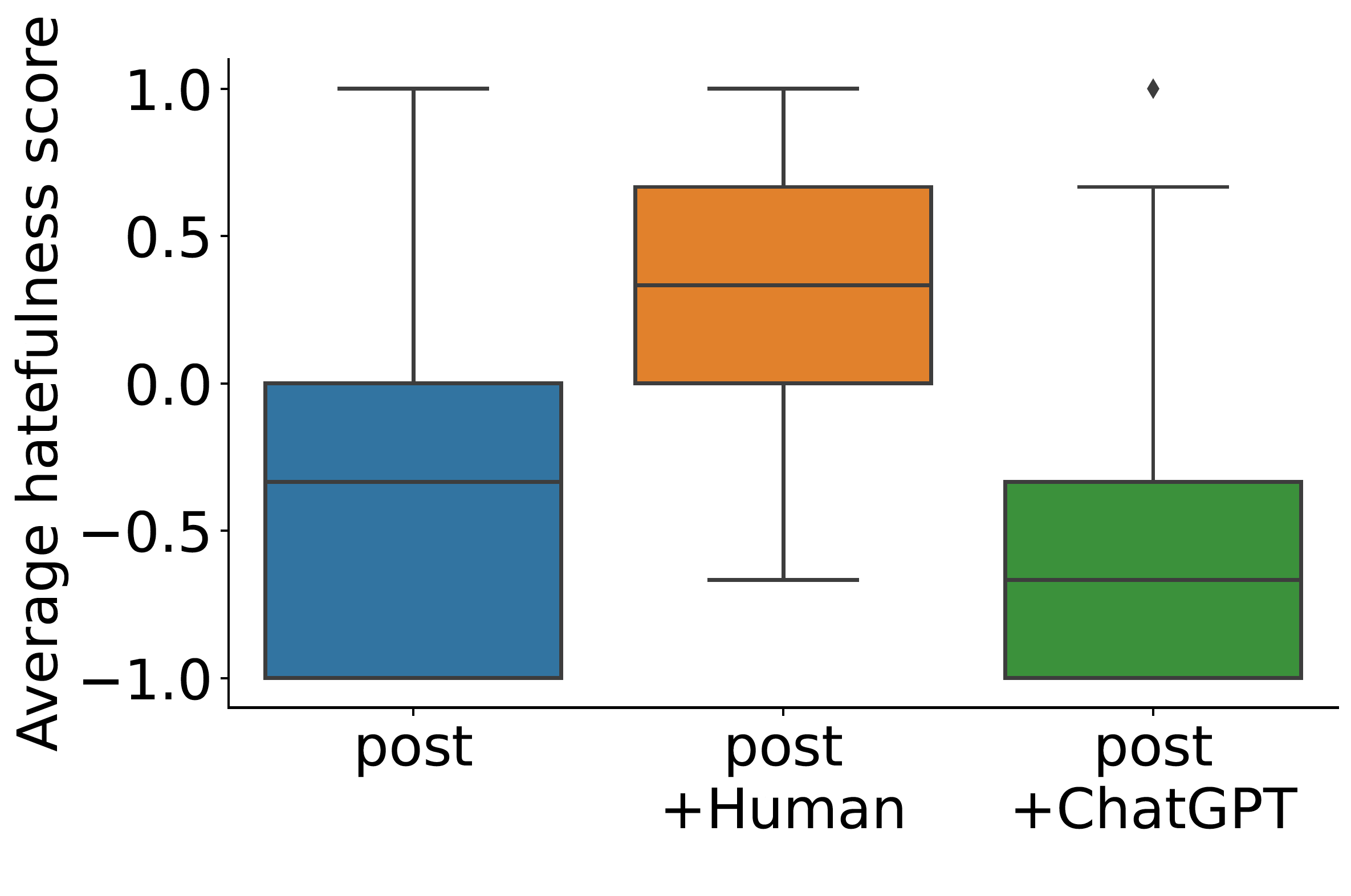}
         \caption{Human evaluation for disagreement between \cite{elsherief-etal-2021-latent} and ChatGPT. We give MTurk workers 1) a post only, 2) a post and human-written NLE (``post+Human''), and 3) a post and ChatGPT-generated NLE (``post+ChatGPT'').}
         \label{fig:disagree}
\end{figure}

There are several interesting observations from the result in Figure~\ref{fig:disagree}. 
First, when only a post is given, we find the average hatefulness score is negative (mean: -0.41), which means lay people (i.e., Mturk workers) are less likely to consider a tweet hateful 
when ChatGPT could not find hate in it. 
Second, the average hatefulness score is negative (mean: -0.52) when a post is given with the ChatGPT-generated NLE. Interestingly, lay people are more likely to find a given tweet non-hateful when ChatGPT explains why the given tweet is non-hateful (mean: -0.52) compared to when only a post is given (mean: -0.41). The difference is statistically significant, confirmed by the t-test (t(316) = -13.75, p$<$0.001). 
Third, the average hatefulness score is positive (mean: 0.29) when a post is given with the human-written NLE. 
This shows that lay people's decisions can be drifted by the additional explanation; however, the lower absolute score indicates that providing human-written NLE is not as convincing as the other two cases. 



Considering the diversity and nuanced nature of the implicit hate speech~\cite{elsherief-etal-2021-latent} and overall toxic behavior online~\cite{kwak2015exploring}, it is not surprising that lay people's decisions can be affected by additional context. 
Nevertheless, what we observe from the above experiment clearly demonstrates the potential of ChatGPT for nuanced, subjective tasks. 
Lay people tend to agree with ChatGPT that those tweets are not likely to be hateful.
However, on the other hand, ChatGPT-generated NLEs \emph{reinforce} the human perception. In a case when ChatGPT's decision is wrong, such capability might be a danger to misleading lay people and potentially constructing wrong labels.

In summary, our answer to RQ1 is \emph{yes}. ChatGPT shows 80\% agreement with \cite{elsherief-etal-2021-latent}. 
Furthermore, for the 20\% disagreement cases, lay people are highly likely to lean toward ChatGPT's classification results. 


\subsection{RQ2: Does ChatGPT generate quality NLEs?}

\begin{figure}[h!]
    \centering
         \includegraphics[width=0.8\columnwidth]{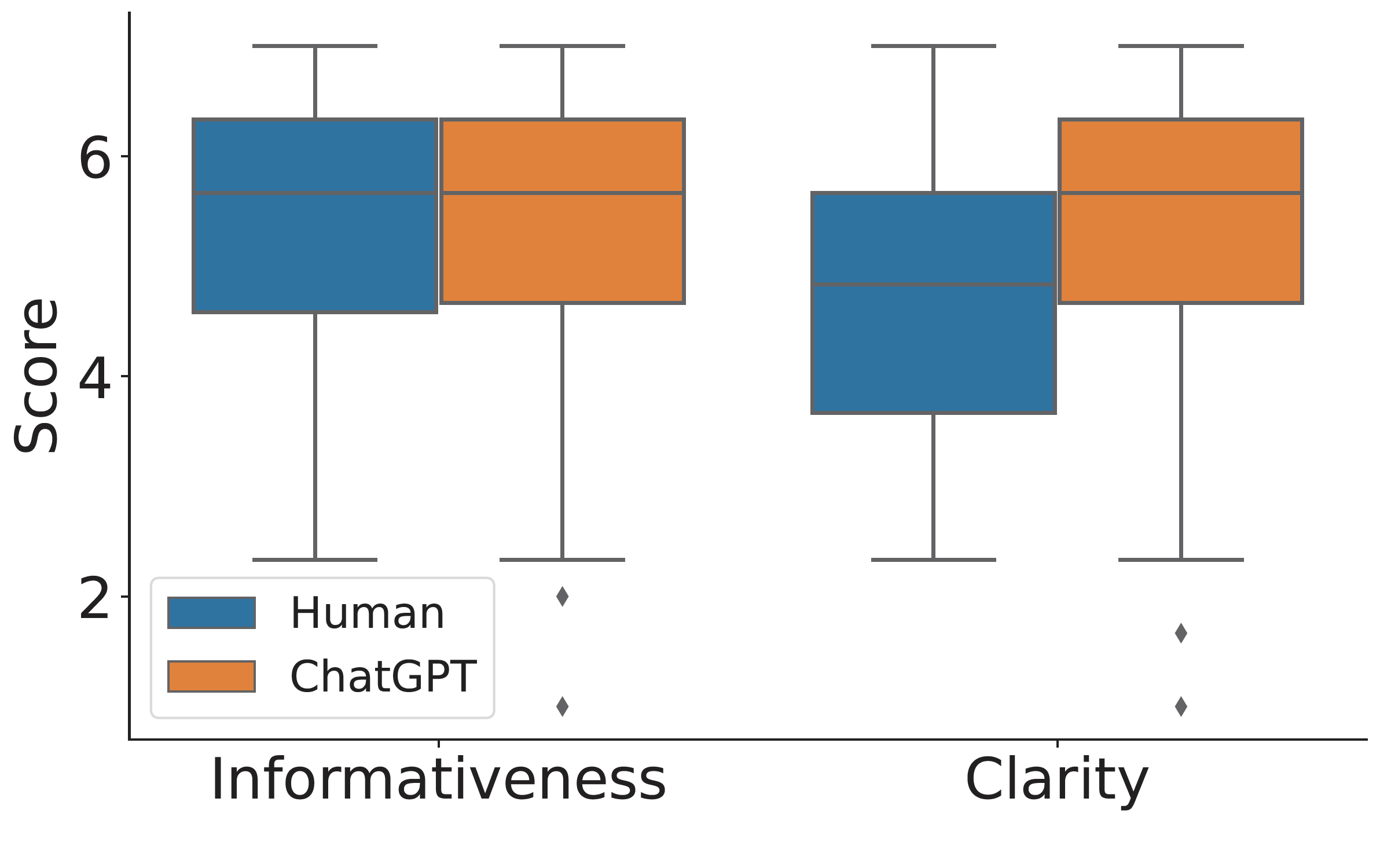}
         \caption{Quality evaluation of human-written NLEs and ChatGPT-generated NLEs}
         \label{fig:informativeness_clarity}
\end{figure}

For the randomly selected 100 instances from our dataset, we collect the Informativeness and Clarity scores as we explained in $\S$3.3. 
In Figure~\ref{fig:informativeness_clarity}, ChatGPT-generated NLEs show significantly higher Clarity scores (mean: 5.39) than human-written NLEs (mean: 4.68). As confirmed by the t-test, the difference is statistically significant (t(197) = -3.73, p < 0.001). 
However, there is no statistically significant difference in Informativeness (t(198) = -0.69, p = 0.49). 

Our findings demonstrate that, for end users, both human-written NLEs and ChatGPT-generated NLEs provide information to help users identify implicit hatefulness. However, ChatGPT provides more comprehensive illustrations so users can easily confirm implicit hatefulness from a given tweet. 

In summary, our answer to RQ2 is also \emph{yes}. ChatGPT generates quality NLEs comparable to human annotators for implicit hate speech. 




\section{Discussions and Conclusion}

In this paper, we examine ChatGPT-generated NLEs for the implicit hateful speech from the perspectives of their accuracy and quality.  

ChatGPT correctly identifies 80\% of the implicit hateful tweets in our experiment setting. The results demonstrate the great potential of ChatGPT as a data annotation tool using a simple prompt design. Further studies can be done to investigate the effect of different prompt designs. 
Furthermore, for those 20\% disagreement cases, our experiments show that ChatGPT's results are more likely to align with lay people's perceptions. Moreover, we also find that the ChatGPT-generated NLEs can reinforce human perception, and they tend to be perceived as clearer than human-written NLEs. 
These results pose an important question.
Since ChatGPT can be convincing, its capability would become a risk of misleading lay people if its decision is wrong. 
This highlights that it would require extra caution when using ChatGPT as a tool for assisting data annotation. 


\section{Ethical Considerations}
Our data collection design via Amazon Mechanical Turk (Mturk) has been approved by Singapore Management University (Approval No.: IRB-22-076-A043(622)). To prevent human annotators from feeling offended or having mental discomfort, we highlight the possible hatefulness in the headline of the Mturk tasks. We also suggest annotators stop labeling anytime they felt uncomfortable and provide the consultant hotlines in our task instructions. 


\bibliographystyle{ACM-Reference-Format}
\bibliography{references}



\end{document}